%% file: acl_latex.tex
\documentclass[11pt]{article}

\usepackage[final]{acl}

\usepackage{times}
\usepackage{latexsym}
\usepackage{xcolor}
\usepackage{graphicx}
\usepackage{booktabs}
\usepackage{multirow}
\usepackage{ragged2e}
\usepackage[many]{tcolorbox}
\usepackage{array}
\usepackage{float}

\usepackage[T1]{fontenc}

\usepackage[utf8]{inputenc}

\usepackage{microtype}

\usepackage{inconsolata}

\usepackage{graphicx}

\title{Quantifying the Statistical Effect of Rubric Modifications on Human-Autorater Agreement}

\author{Jessica Huynh \\
  Carnegie Mellon University \\
  \texttt{jhuynh@cs.cmu.edu} \\\And
  Alfredo Gomez \\
  Carnegie Mellon University \\
  \texttt{alfredo3@cs.cmu.edu} \\\And
  Athiya Deviyani \\
  Carnegie Mellon University \\
  \texttt{adeviyan@cs.cmu.edu} \\\AND
  Renee Shelby \\
  Google Research \\
  \texttt{reneeshelby@google.com} \\\And
  Jeffrey P. Bigham \\
  Carnegie Mellon University \\
  \texttt{jbigham@cmu.edu} \\\And
  Fernando Diaz \\
  Carnegie Mellon University \\
  \texttt{diazf@cmu.edu} \\}

\begin{document}
\maketitle
\begin{abstract}
Autoraters, also referred to as LLM-as-judges, are increasingly used for evaluation and automated content moderation. However, there is limited statistical analysis of how modifications in a rubric presented to both humans and autoraters affect their score agreement. Rubrics that ask for an overall or \emph{holistic} judgment - for example, rating the ``quality'' of an essay - may be inconsistently interpreted due to the complexity or subjectivity of the criteria. Conversely, rubrics can ask for \emph{analytic} judgments, which decompose assessment criteria - for example, ``quality'' into ``fluency'' and ``organization''. While these rubrics can be edited to improve the individual accuracy of both human and automated scoring, this approach may result in disagreement between the two scores, or with the associated holistic judgment. Designing and deploying reliable autoraters requires understanding not just the relationship between human and autorater annotations but how that relationship changes as holistic or analytic judgments are elicited. The results indicate that rubric edits providing representative examples and additional context, and reducing positional bias in the rubric increased human-autorater agreement, while higher rubric complexity and conservative aggregation methods tended to decrease it. The findings from the automatic essay scoring and instruction-following evaluation domains suggest that practitioners should carefully analyze domain- and rubric-specific performance to move towards higher human-autorater agreement. 
\end{abstract}

\input{sections/introduction}
\input{sections/related_work}
\input{sections/methods}

\input{sections/results}
\input{sections/conclusion}

\input{sections/appendix/limitations}

\bibliography{custom}

\appendix

\input{sections/appendix}

\end{document}

%% file: sections/introduction.tex
\section{Introduction}
\label{sec:intro}

\begin{figure}[hbt!]
    \includegraphics[width=\linewidth]{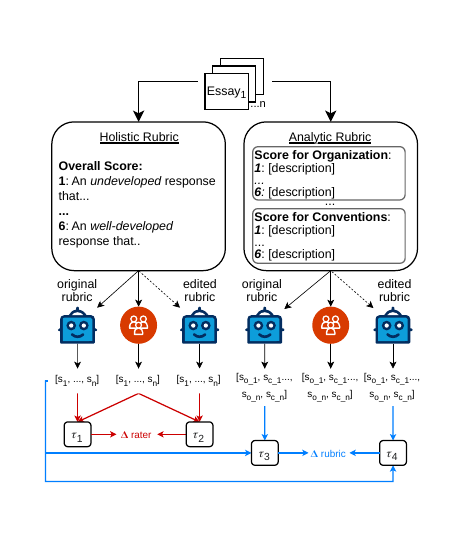}
    \vspace{-1cm}
    \caption{This diagram provides a walkthrough of the experimental setup as shown for automatic essay scoring.
    It represents comparisons made between human-autorater agreements $\tau$ across holistic rubrics (left), in which all criteria are applied together in a single overall judgment, or analytic rubrics (right), in which criteria are evaluated separately, resulting in multiple scores. The original rubrics are given to a human and an autorater, while edited rubrics are given to autoraters only. Arrows in bold between $\tau_1$, $\tau_2$ and $\tau_3$, $\tau_4$ represent comparisons for which statistical significance can be calculated. $\Delta\textnormal{rater}$ represents comparisons where the type of rater is changed while the type of rubric remains constant, while $\Delta\textnormal{rubric}$ represents comparisons where the type of rubric is changed and the type of rater remains constant.}
    \label{fig:experiment_design}
\end{figure}

Autoraters, or LLM-as-judges, have been used as an alternative to human annotation due to their scalability, cost, and time effectiveness. Their effectiveness is typically validated through agreement with human annotation. 
Borrowing from education literature, a \textit{rubric} is defined as having ``\textit{coherent sets of criteria}" and ``\textit{descriptions of levels of performance for these criteria}" \cite{brookhart2013create}. This describes the scoring guidelines and instructions provided to any rater, whether human or automated, which is also referred to as part of a prompt as described in autorater literature.
Ideally, both human raters and autoraters would receive equivalent evaluation rubrics that accurately measure the same construct with reliable certainty.
However, equivalence does not necessitate identical presentation. 
\citet{Wu_Quinn_2017} show that expert and non-expert human raters may require different levels of instruction specificity, for instance, specifying tools and providing concrete examples improves accuracy specifically when raters lack task-relevant knowledge. 

In addition to identical presentation across different types of raters, humans and autoraters are sensitive to variations in instruction presentations within the same type of rater. For humans, this sensitivity includes the interpretation of the task during crowdsourcing \citep{10.1145/2818048.2820016} and instruction specification, where increasing instruction specification increases accuracy on the task \citep{Wu_Quinn_2017}. For autoraters, position bias (where the position of the evaluated text within the prompt will influence the autorater evaluation) and verbosity bias (where autoraters prefer more verbose texts) are only a few of the known sensitivities \citep{10.5555/3666122.3668142}. Autoraters are also sensitive to rubric variations such as formatting choices \cite{sclar2024quantifying} and example ordering \cite{lu-etal-2022-fantastically}. Another example of instruction presentation involves decomposing criteria into multiple sub-criteria. Previous work has successfully used decomposition to improve LLM performance, whether by having autoraters decompose evaluation criteria into sub-tasks \citep{saha-etal-2024-branch} or by having humans decompose complex questions into simpler sub-questions for models
\citep{patel-etal-2022-question}.

Understanding whether rubric modifications produce statistically significant shifts in agreement is essential for practitioners who aim to deploy autoraters as evaluation tools. This work studies human-autorater agreement in two domains: automatic essay scoring (AES) and instruction-following (IF). We examine statistically how rubric presentation and broader rubric modifications affect human-autorater agreement on subjective evaluation tasks, as well as empirically examining whether decomposing general holistic judgments that ask for a single high-level criteria into more granular sub-criteria, originally designed for human evaluators, can be an effective approach for improving human-judge agreement. The findings indicate that instructions optimized for autoraters tend to improve agreement with human ratings when autoraters receive machine-optimized instructions and humans receive the original set of instructions.
Conversely, giving autoraters simpler prompts does not guarantee higher agreement with human ratings.
These results indicate that 1) rubric edits providing representative examples along with contextual information increased human-autorater agreement as well as autorater self-agreement, 2) higher criterion complexity and conservative aggregation methods tended to decrease human-autorater agreement, 3) reducing confirmation bias tends to significantly increase human-autorater agreement and 4) high human inter-rater agreement leads to significantly higher human-autorater agreement.
These findings from the automatic essay scoring and instruction-following evaluation domains suggest that practitioners should carefully analyze domain-specific performance and modify rubrics to move towards high human-autorater agreement. 

%% file: sections/related_work.tex
\section{Related Work}


\subsection{Rubrics in Context}

Expanding upon Section~\ref{sec:intro}, rubrics consist of both the criteria, which are the components of the overall evaluation, and the descriptions for the criteria. 

\textbf{Decomposition level} describes criteria presentation, and refers to whether prompts are \textit{holistic}, where ``all criteria are [applied] at the same time'', or \textit{analytic}, where ``work [is described] on each criterion separately'', \citep{brookhart2013create}. 

\textbf{Generality level} details the descriptions for the criteria, and refers to whether prompts are \textit{general} or \textit{task-specific} (i.e., evaluation prompts that [can$\backslash$cannot] also be used for other tasks) \citep{brookhart2013create}. Autorater evaluation prompts can be viewed through this lens - for example, a holistic autorater evaluation prompt may ask for a single overall judgment whereas an analytic prompt would decompose the evaluation into criterion to be evaluated separately.


\textbf{Prompt complexity}, as used in this work, refers to the cognitive demands placed on a rater during evaluation. Prior work on task complexity and cognitive demand has identified the number of \textit{paths}, or components, that must be considered simultaneously \citep{8163de9b-fb12-3776-8b6f-4c14075da748}, the degree of element interactivity (i.e., the extent to which components must be processed together rather than independently) \citep{sweller_element_2010} and ambiguity resulting in communication failure \citep{8163de9b-fb12-3776-8b6f-4c14075da748} as characteristics of cognitive load. Thus in our work, we consider the number of criteria evaluated, the degree to which score-level descriptions are interrelated, and the extent to which the rater must resolve ambiguity across sub-criteria. 

\subsection{LLM-as-judges}

The autorater (LLM-as-a-judge) paradigm has received substantial recent attention. Several past works demonstrated that LLMs can produce evaluations consistent with human experts. \citep{chiang_can_2023, li-etal-2025-generation}. However, research also shown that autoraters are sensitive to prompt variations, with different instructions leading to substantially different performances \cite{mizrahi-etal-2024-state} and varying quality, necessitating statistical procedures to justify replacing human raters \cite{calderon-etal-2025-alternative}. Several recent works have studied how prompt modifications affect autorater performance. \citep{sclar2024quantifying} demonstrated that prompt formatting choices (e.g., separator characters, whitespace) can significantly affect task accuracy, though these effects weakly correlate across models. \citep{lu-etal-2022-fantastically} showed that example ordering in few-shot prompts substantially impacts performance on classification tasks. Our work extends this literature by using statistical procedures where possible to test how rubric modification choices, such as decomposition level, example selection, and aggregation methods affect agreement with human judgment on tasks where even trained human raters often disagree. This work also empirically examines whether simpler rubrics, often assumed to reduce cognitive load, actually improve human-autorater agreement, in domains where there is no single correct answer, contrasting with the classification and multiple-choice tasks used in prior work.

\subsection{Automatic Essay Scoring (AES) and Instruction-Following (IF)}

Recent work has explored using autoraters for AES, employing various personas, including a ``virtual evaluator with expertise in English composition'' \citep{10.1145/3706468.3706507}, a ``helpful pattern-following assistant'' \citep{mansour-etal-2024-large}, and an ``English essay writing test evaluation committee'' or ``English teacher'' \citep{lee-etal-2024-unleashing}. However, these personas may be misaligned with the original human raters; for example, scoring guidelines for a portion of the dataset from \citet{asap-aes} explicitly state that raters should not be teachers. The number of few-shot examples provided to the autorater also varies, \citet{10.1145/3706468.3706507} select the closest three examples with calculated embeddings, while \citet{kundu2024large} chose one essay that scored highly and one essay that scored poorly. 

While most studies utilize the rubrics given by the AES datasets, \citet{lee-etal-2024-unleashing} automatically decomposed the original essay scoring rubric into sub-criteria and performed a modified average aggregation on the sub-criteria, which on average performs better than using a single score from zero-shot prompting. However, \citet{10.1145/3706468.3706507} demonstrated significant improvements by fine-tuning GPT-3.5-turbo and Llama3-8B compared to using GPT-4 with rubrics and few-shot examples. These previous works also may request explanations for scores from autoraters to mimic CoT, aiming for more accurate explanations and ratings. 
Studies on analytic rubrics primarily focus on cross-prompt scoring \citep{chen-li-2023-pmaes}. The experimental setup in this work follows the rigor of prior work by 
studying multiple autoraters, using given rubrics, and performing various rubric edits. However, the hypotheses examined extend prior findings on edited rubrics and further investigate rubric components across various essay scoring rubrics.

Research on instruction-following in LLMs has led to two primary approaches for improving and evaluating LLMs, instruction tuning and alignment tuning. Prior work has found that larger models tend to follow instructions more accurately, though this relationship is not strictly linear \citep{ouyang2022training}. \citet{honovich-etal-2023-instruction} proposed allowing LLMs to write instructions based on only seeing examples of a task, although this is still less accurate than human-written instructions.



%% file: sections/methods.tex

\section{Experiment Setup}


Our hypotheses focus on the criteria presentation of rubrics for autoraters:
(1) \textbf{Edited} prompts will improve autorater alignment with human ratings over the original prompts (humans and autoraters may require different prompts due to the way that they process information); and 
(2) \textbf{Analytic} rubrics will improve autorater alignment with human ratings over \textbf{holistic} rubrics (the decomposition of holistic constructs into discrete components within analytic rubrics is expected to improve autorater alignment by simplifying the evaluation task).

\subsection{Experiments}
\label{sec:experiments}

This work examines four different scores: human ratings on holistic ($\textnormal{H}_{H}$) and analytic prompts ($\textnormal{H}_{A}$), and autorater ratings on holistic ($\textnormal{LLM}_{H}$) and analytic prompts ($\textnormal{LLM}_{A}$). These scores facilitate two comparisons: \textbf{$\Delta$Rater}, which assesses human-autorater agreement when using the same type of rubric, \textbf{$\Delta$Rubric}, which investigates the impact of varying the type rubric while keeping the type of rater constant, revealing how different rubrics influence ratings even when measuring the same criteria. These comparisons are depicted in Figure \ref{fig:experiment_design}. Additional analysis on \textbf{$\Delta$Rater+Rubric}, varying both the type of rubric and type of rater, is explored in Appendix~\ref{sec:other_obs} for completion.

Additionally, this study compares two approaches: presenting the autorater with the original human rubric versus a modified rubric, or \textbf{edited rubric}, designed to enhance agreement. Modifications include adding additional context (if available), incorporating examples, and reducing positional bias. Prior work \citep{mansour-etal-2024-large} studied adding a rubric and then adding an example into the prompt, and found that ChatGPT\footnote{\label{note}\url{https://platform.openai.com/docs/models}} benefited from having rubrics and examples, whereas Llama-2-13b-chat-hf\footnote{\url{https://huggingface.co/meta-llama/Llama-2-13b-chat-hf}} did not for some cases. This work uses several modification conditions. First, the original holistic prompts for the AES task include a large number of examples (10-18) - edited holistic prompts reduces the examples to a representative set of three examples (\textbf{3ex}; high, medium, and low scores). This serves two purposes: (1) it creates a more direct comparison with the \textbf{edited} analytic prompts, which also use three examples, and (2) it avoids an excessively long prompt for the model. Second, the analytic rubric prompt formats of presenting all sub-criteria in a single API call (batch), in separate API calls to mitigate potential positional bias (separate), and a combination of enhancements such as incorporating additional context, the "3ex" set (as the original analytic rubrics had no examples), and the "separate" API call strategy (edited) are tested. Prompts are provided in Appendix~\ref{sec:appendix_B}. 

\section{Methods and Materials}

\subsection{Datasets}
\label{sec:datasets}

Each hypothesis is tested across the domains of AES and IF to determine whether any observed effect is context-dependent. Human annotations are provided by each dataset.

\paragraph{Automatic Essay Scoring (AES).} Automatic essay scoring is a well-established field, with the Automated Student Assessment Prize (ASAP) as a primary dataset \citep{asap-aes}. ASAP contains over 21,000 essays over 8 essay prompts, each with a holistic rubric used by trained human raters. To provide further diversity within AES, this study uses essay prompts 1, 4, and 6, which represent diverse essay prompt types (argumentative, source-dependent, source-dependent) and rubric types (general, general, task-specific) shown in Appendix Table~\ref{tab:asap_prompts}. The essay prompts also provide single holistic scores, and integer scores from 1 to 5 (essay prompt 6's scores are condensed to integers). 

The ASAP++ dataset \citep{mathias-bhattacharyya-2018-asap} provides corresponding analytic rubrics for these essay prompts. Except for the holistic rubric for essay prompt 6, all other rubrics are general. It is important to note while both ASAP and ASAP++ evaluate essay quality, the specific sub-criteria for essay quality differ (ex. ``audience awareness'' is mentioned in ASAP but not ASAP++ and ``conventions'' is mentioned in ASAP++ but ASAP explicitly asks raters to not consider this). This discrepancy may lead to imperfect agreement between ASAP and combined ASAP++ ratings. Additionally, the ASAP++ analytic rubrics do not include examples, so three examples per essay prompt were selected from ASAP++ and excluded from analysis. These examples represent high, medium, and low scores across all analytic sub-criteria (ex. a high scoring essay contains the highest scores for all analytic sub-criteria).

\paragraph{Instruction-following (IF).} 
Compared to AES, 
instruction-following work has decomposed instructions into a series of questions that isolate individual criteria, but with no evaluation differences in the rubric. The InfoBench \citep{qin-etal-2024-infobench} dataset consists of both easy and hard instructions, along with outputs from five popular LLMs. Outputs are holistically scored on a scale of one to five with a general rubric, and analytically scored with binary yes/no responses to decomposed instructions with task-specific rubrics. In the dataset, each output was annotated with both scores by three expert annotators using provided rubrics. Expert annotators annotated 50 instructions with 5 generations each. The same example selection methodology used for AES was used to select input, output, instruction tuples from the remaining dataset that were not annotated by the expert annotators, since neither rubric in IF provides examples. The decomposed automatic instruction annotations released by the InfoBench authors served as a starting point to explore the tuples. The resulting example outputs were generated from three models: GPT-3.5-turbo\footref{note}, Alpaca-7b~\citep{taori2023alpaca}, and GPT-4\footref{note}\citep{openai2024gpt4technicalreport}), and represent high, medium, and low instruction-following ratios (100\%, 50\%, 0\%). This is translated into holistic scores of 5, 3, and 1, respectively, which were validated by the authors.

We select AES and IF for complementary purposes. The AES dataset provides validity through its multiple rubric variations (examples) all measuring the same underlying construct, essay quality, allowing us to observe the different effects of different operationalizations of the same evaluation task. Although using ASAP and ASAP++ simultaneously may result in a noisy comparison, it provides preliminary evidence for the behavior. Conversely the InfoBench dataset offers a direct 1:1 comparison holistic and analytic rubrics that measure identical criteria. Together, these datasets enable us to distinguish between effects that are domain specific versus those that generalize across evaluation contexts/tasks.

\subsection{Autoraters}
Experiments were conducted using gpt-4o-2024-11-20\footref{note}\citep{openai2024gpt4ocard} as the autorater, which has demonstrated high alignment with human evaluations across various tasks \citep{chan:chateval, zhou:sotopia}, and Llama-3.1-70B-Instruct \citep{grattafiori2024llama}. 


Autorater ratings are calculated using a probability weighting scheme similar to \citet{huynh2023understanding}, given by the equation: $r = \sum_{i=0}^{n}\frac{{p_i}}{\sum_{j=0}^{n} {p_j}} *s_i$, where $r$ is the final rating from the autorater, $n$ represents the number of possible scores given by the rubric, $p_i$ represents the exponential of the log probability score given by the autorater, and $s_i$ represents the integer score outputted by the autorater that corresponds to the log probability. In IF for analytic prompts, $s_i$ is represented by 1 for ``yes'' and 0 for ``no''.

\begin{table*}[!ht]
    \small
    \centering
    \begin{tabular}{|w{c}{0.05cm}|w{c}{0.3cm}||c|c||c|c|c||c|c||c|c|c|}
    \hline
        & & \multicolumn{5}{c||}{GPT-4o} & \multicolumn{5}{c|}{Llama}\\
        \hline
        & & \multicolumn{10}{c|}{$\Delta\textbf{rater}$} \\
        \hline
        & & \multicolumn{2}{c||}{$\textnormal{H}_{H} \to \textnormal{LLM}_H$}  & \multicolumn{3}{c||}{$\textnormal{H}_{A} \to \textnormal{LLM}_A$} & \multicolumn{2}{c||}{$\textnormal{H}_{H} \to \textnormal{LLM}_H$}  & \multicolumn{3}{c|}{$\textnormal{H}_{A} \to \textnormal{LLM}_A$}\\
        \hline
        P & C. & Full & 3ex & Sep. & Bat. & Edited& Full & 3ex & Sep. & Bat. & Edited \\
        \hline
        \multirow{6}{*}{1}& H. & $0.437^{\dagger}$ & $0.387$ & - & - & - & $0.576$ & $0.571$ & - & - & -\\
        \cline{2-12}
        & Ide. & -& -& $0.474^{\star}$ & $0.464$ & $0.552^{s,b\uparrow}$ & -& -& $0.552$ & $0.559$ & $0.552$\\
        \cline{2-12}
        & Or. & -& -& $0.497^{\star}$ & $0.448$ & $0.544^{s,b\uparrow}$ & -& -&$0.547^{\star}$ & $0.524$ & $0.529^{s\downarrow}$\\
        \cline{2-12}
        & WC & -& -& $0.452^{\star}$ & $0.439$ & $0.554^{s,b\uparrow}$ & -& -& $0.547^{\star}$ & $0.527$ & $0.507^{s\downarrow}$\\
        \cline{2-12}
        & SF & -& -& $0.459$ & $0.455$ & $0.553^{s,b\uparrow}$ & -& -& $0.559^{\star}$ & $0.512$ & $0.523^{s\downarrow}$\\
        \cline{2-12}
        & Cv. & -& -& $0.362$ & $0.388^{\star}$ & $0.472^{s,b\uparrow}$ & -& -& $0.491$& $0.477$ & $0.487$\\
        \hline\hline
        \multirow{5}{*}{4}& H. & $0.695^{\dagger}$ & $0.687$ & - & - & - & $0.700$ & $0.699$ & - & - & -\\
        \cline{2-12}
        & Ct. & -& -& $0.673$ & $0.696^{\star}$ & $0.698^{s\uparrow}$& -& -& $0.698^{\star}$ & $0.693$ & $0.701$\\
        \cline{2-12}
        & PA & -& -& $0.662$ & $0.677^{\star}$ & $0.680^{s\uparrow}$& -& -& $0.683^{\star}$ & $0.665$ & $0.682^{b\uparrow}$\\
        \cline{2-12}
        & La. & -& -& $0.627^{\star}$ & $0.595$ & $0.639^{s,b\uparrow}$& -& -& $0.630^{\star}$ & $0.573$ & $0.621^{b\uparrow}$\\
        \cline{2-12}
        & Na. & -& -& $0.669^{\star}$ & $0.652$ & $0.682^{s,b\uparrow}$& -& -& $0.674^{\star}$ & $0.642$ & $0.669^{b\uparrow}$\\
        \hline\hline
        \multirow{6}{*}{6} & H. & $0.629$ & $0.644^{\dagger}$ & - & - & -& $0.666$ & $0.680^{\dagger}$ & - & - & -\\
        \cline{2-12}
        & Ct. & -& -& $0.610$ & $0.652^{\star}$ & $0.676^{s,b\uparrow}$& -& -& $0.680^{\star}$ & $0.648$ & $0.694^{s,b\uparrow}$\\
        \cline{2-12}
        & PA & -& -& $0.605$ & $0.608$ & $0.666^{s,b\uparrow}$ & -& -& $0.652^{\star}$ & $0.601$ & $0.668^{s,b\uparrow}$\\
        \cline{2-12}
        & La. & -& -& $0.524^{\star}$ & $0.510$ & $0.546^{s,b\uparrow}$& -& -& $0.542^{\star}$ & $0.496$ & $0.547^{b\uparrow}$\\
        \cline{2-12}
        & Na. & -& -& $0.562$ & $0.557$ & $0.590^{s,b\uparrow}$& -& -& $0.579^{\star}$ & $0.547$ & $0.589^{b\uparrow}$\\
        \hline\hline
        & & \multicolumn{10}{c|}{$\Delta\textbf{rubric}$} \\
        \hline
        & & \multicolumn{2}{c||}{$\textnormal{H}_{H} \to \textnormal{H}_A$}& \multicolumn{3}{c||}{$\textnormal{LLM}_H \to \textnormal{LLM}_A$} & \multicolumn{2}{c||}{$\textnormal{H}_{H} \to \textnormal{H}_A$}& \multicolumn{3}{c|}{$\textnormal{LLM}_H \to \textnormal{LLM}_A$}\\
        \hline
        P & C. & Full & 3ex & Sep. & Bat. & Edited& Full & 3ex & Sep. & Bat. & Edited \\
        \hline
        \multirow{2}{*}{1} & full &$0.591$ &- & $0.756$& $0.750$& $0.839^{s,b\uparrow}$ &- &- & $0.685^{\star}$ & $0.671$& $0.711^{s,b\uparrow}$\\
        \cline{2-12}
        & 3ex &- &- & $0.792$& $0.789$& $0.838^{s,b\uparrow}$ &- &- & $0.736^{\star}$& $0.713$& $0.768^{s,b\uparrow}$\\
        \hline
        \multirow{2}{*}{4} & full &$0.656$ &- &$0.838^{\star}$ & $0.817$ & $0.876^{s,b\uparrow}$ &- &- & $0.854^{\star}$& $0.780$ & $0.869^{s,b\uparrow}$\\
        \cline{2-12}
        & 3ex&- &- & $0.847^{\star}$& $0.823$& $0.880^{s,b\uparrow}$ &- &- & $0.861^{\star}$& $0.784$& $0.877^{s,b\uparrow}$\\
        \hline
        \multirow{2}{*}{6} & full &$0.681$ &- & $0.781^{\star}$& $0.773$& $0.842^{s,b\uparrow}$ & -&- & $0.803^{\star}$& $0.748$& $0.864^{s,b\uparrow}$\\
        \cline{2-12}
        & 3ex&- &- & $0.804$& $0.801$& $0.857^{s,b\uparrow}$ &- &- & $0.813^{\star}$& $0.751$& $0.868^{s,b\uparrow}$\\
        \hline
    \end{tabular}
    \caption{Kendall's $\tau$ results on AES with GPT-4o and Llama for $\Delta$rater. P. indicates the essay prompt, and C. indicates what ratings are being compared, with ideas, organization, word choice, sentence fluency, and conventions compared for prompt 1, and content, prompt adherence, language, and narrativity compared for prompts 4 and 6. $\tau$ is calculated with singular numerical values for $\Delta$rater and calculated through Pareto dominance comparison for preferences for $\Delta$rubric. Significance tests between separate (sep.), batch (bat.), and edited prompts are performed, where ${^s}$ and ${^b}$ in the edited prompt column represents significant differences with separate and batch prompts respectively. $\dagger$ is indicated next to comparisons that are significantly larger within holistic prompts. $\star$ is indicated next to comparisons that are significantly larger between separate and batch comparisons. $\uparrow$ and $\downarrow$ represent that the $\tau$ value for edited prompts is significantly larger or smaller respectively with the separate $s$ or batch $b$ prompts' $\tau$. The lack of any dagger, star, or arrow denotes no statistical significance. $\textnormal{H}$ is a shortened form for $\textnormal{Human}$. }
    \label{tab:asap_results}
\end{table*}

\begin{table*}[h!]
    \centering
    \begin{tabular}{|c|w{c}{0.3cm}||c|c||c|c|c||c|c||c|c|c|}
    \hline
        & & \multicolumn{5}{c||}{GPT-4o} & \multicolumn{5}{c|}{Llama}\\
        \hline
        \multirow{3}{*}{\rotatebox[origin=c]{90}{$\Delta$rater}} & & \multicolumn{2}{c||}{Holistic} & \multicolumn{3}{c||}{Analytic} & \multicolumn{2}{c||}{Holistic} & \multicolumn{3}{c|}{Analytic}\\
        \cline{3-12}
         & & 0ex & 3ex & Sep. & Bat. & Edited & 0ex & 3ex& Sep. & Bat. & Edited\\
        \cline{3-12}
         & & $0.536$& $0.585$& $0.464^{\star}$& $0.167$&$0.471^{b\uparrow}$ &$0.470$& $0.578$& $0.445^{\star}$& $0.166$&$0.426^{b\uparrow}$\\
        \hline\hline
             \multirow{3}{*}{\rotatebox[origin=c]{90}{$\Delta$rubric}} & Ex. & \multicolumn{2}{c||}{$\textnormal{H}_{H} \to \textnormal{H}_A$}& \multicolumn{3}{c||}{$\textnormal{LLM}_H \to \textnormal{LLM}_A$} & \multicolumn{2}{c||}{$\textnormal{H}_{H} \to \textnormal{H}_A$}& \multicolumn{3}{c|}{$\textnormal{LLM}_H \to \textnormal{LLM}_A$}\\
        \cline{2-12}
        & 0ex & $0.534$&-& $0.640^{\star}$& $0.455$& $0.640^{b\uparrow}$ & $0.551$&- & $0.531^{\star}$& $0.285$& $0.551^{b\uparrow}$\\
        \cline{2-12}
         & 3ex &-&-& $0.623^{\star}$& $0.484$& $0.617 ^{b \uparrow}$ &- &- & $0.566^{\star}$& $0.326$& $0.623^{b\uparrow}$\\
        
        \hline
    \end{tabular}
    \caption{Kendall's $\tau$ results on IF with GPT-4o and Llama. $\Delta$rubric is calculated through instruction following ratio comparison for preferences. All other calculations and significance follow the methodology of Table~\ref{tab:asap_results}.}
    \label{tab:if_results}
\end{table*}

\subsection{Agreement Calculation}

While direct numerical comparison (e.g., correlation between raw scores) is often calculated through agreement metrics such as Cohen's $\kappa$, autoraters and human have been shown to use the same scales differently, with autoraters often exhibiting compressed or shifted score distributions relative to humans~\cite{kundu2024large}. Preference-based evaluation (e.g., A/B) is more robust to such shifts, as it captures ordinal relationships rather than numerical ones; however, such methods do not scale with large datasets. Addressing both issues, we use Kendall's $\tau$ with tie consideration (calculated using the SciPy \citep{virtanen_scipy_2020} implementation of \citep{df984cd0-5ddb-387f-ae81-1c7b41e92c35}), which operates on the existing numerical scores from the data while evaluating the scores through pairwise ordinal comparisons.

Each domain employs a different method for aggregating analytic rubric scores when comparing them to holistic rubric scores to calculate concordant pairs for Kendall's $\tau$. In AES, Pareto dominance is used, where essay A is considered better than essay B if all sub-criteria scores of A are at least tied to those of B, with at least one sub-criteria score from A being higher. Pareto dominance is used as a conservative aggregation method to ensure that one essay would truly be better than another without knowing about how the sub-criteria were factored into a holistic rating. In IF, response A is considered to follow instructions better than response B if A has a higher ratio of correctly followed instructions.

Pareto dominance is chosen as a conservative aggregation method precisely because the sub-criteria in ASAP and ASAP++ do not perfectly overlap, as noted in \ref{sec:datasets}. Rather than imposing an arbitrary weighting scheme or score cutoff to determine which essay is better, Pareto dominance avoids assuming any particular trade-off between criteria, thus providing a reliable lower bound on the present agreement.

The confidence interval around the correlations is calculated using bootstrapping \citep{efron1992bootstrap} with 1000 samples. The difference between two conditions is calculated, and the 95\% confidence interval is determined by sampling the 25th and 975th sorted values. For comparisons involving three conditions, the confidence interval is adjusted using Bonferroni correction, with the interval bounds set as the average of the 8th and 9th values and the average of the 991st and 992nd values. The compared conditions are considered significantly different if 0 does not fall within the interval.

%% file: sections/results.tex
\section{Results}

\subsection{Edited Rubrics}

When editing rubrics for autoraters to improve human-autorater agreement, it is important to provide examples and context and remove confirmation bias from analytic rubrics.

\textbf{Using GPT with edited analytic AES rubrics mostly significantly improves agreement with humans.} In Table~\ref{tab:asap_results} under $\Delta\textnormal{rater}$ under the ``edited" column, adding examples and context in the edited rubric always improved human-autorater agreement, significantly in the majority of cases, when using GPT, indicated by $s,b\uparrow$. These improvements range from 0.696 (batch prompt 4 content) to 0.698 (edited prompt 4 content) for the smaller, non-significant improvements, to 0.439 (batch prompt 1 word choice) to 0.554 (edited prompt 1 word choice), for the larger, significant improvements. While the analytic rubrics lacked explanations for the examples that the holistic AES rubrics provided, this statistically significant improvement suggests that the examples and context still provide necessary grounding for autorater scores.

\textbf{Edited analytic AES rubrics significantly improves both models' self-alignment but not with IF rubrics.} In addition to improving human-autorater alignment, adding examples and context in the edited rubric always significantly improves GPT and Llama's self-agreement with their scores on the original holistic rubric, although these two rubrics have inherent differences, seen in Table~\ref{tab:asap_results} under $\Delta\textnormal{rubric}$, indicated by $s,b\uparrow$. Interestingly, these agreements surpass human-human agreement with the two types of unedited rubrics (0.591, 0.656, 0.681 compared to 0.750, 0.817, 0.773 for GPT, and 0.671, 0.780, and 0.748 for Llama). This could be because the same autorater is used with both rubrics, whereas human ratings are from different individuals. This suggests that while the prompts superficially evaluate different criteria, the autorater converges to a consistent understanding of essay quality across both rubrics, which aligns with the overall rating objective, a convergence not observed to the same degree in human ratings.

However, adding examples of human analytic scores does not necessarily increase either autoraters' self-alignment on IF. Adding examples decreased $\tau$ with GPT-4o ($\tau$ from 0.640 to 0.640, and 0.623 to 0.617 between separate and edited rubrics in Table~\ref{tab:if_results}), but increased $\tau$ with Llama (0.531 to 0.551 and 0.566 to 0.623 respectively). This suggests that autoraters in this case do not achieve a unified understanding of instruction following across prompts. Despite this, the ratio aggregation method aligns more closely with the autorater's internal reasoning about output instruction adherence. With edited prompts, GPT-4o's alignment (but not Llama's) exceeded human alignment on both prompts, even though the human expert annotators across both rubrics remained consistent.

\textbf{Reducing confirmation bias in all analytic rubrics provides mostly significant improvement for both models.} 
Prior work has highlighted confirmation bias \citep{doi:10.1518/001872008X354183}, in which conceived judgment is reinforced as the task progresses. This bias could affect analytic rubrics, leading raters to assign low scores across all criteria if they initially believe the piece of text is of poor quality. \citet{lee-etal-2024-unleashing} has shown that using separate conversations to rate each sub-criteria tends to increase agreement over using a single conversation to rate all sub-criteria. However, this bias has not been statistically assessed for autoraters, and may affect autoraters differently than humans. 

Across the analytic rubrics on both tasks, the majority of comparisons show separate rubrics significantly outperforming the batched rubrics. With Llama on AES and both models on IF, edited rubrics do not consistently outperform separate rubrics. This suggests the separation of the individual criteria is more important than the examples for Llama. The scoring rubric itself may contribute to the inconsistency for GPT-4o between AES and IF. In AES, example essays answered the same writing prompt as the evaluated essay, and were accompanied by the same scoring rubric during evaluation. Conversely, in IF, the examples did not correspond to the same decomposed questions being scored. The rubrics in IF are task-specific and vary significantly across instructions. 

In addition, there is a trend of $\tau$ decreasing and then increasing based on criteria order in prompts 4 and 6, observed with both GPT-4o (0.696, 0.677, 0.595, 0.652 and 0.652, 0.608, 0.510, 0.557 respectively) and Llama (0.693, 0.665, 0.573, 0.642 and 0.648, 0.601, 0.496, 0.547 respectively). This suggests potential differences in bias between humans and autoraters. This may also suggest that autoraters struggle with rating word choice or language. Qualitative analysis of model outputs for instruction following reveal that batch ratings consistently demonstrated a bias with probability distributions heavily skewed towards ``yes" responses when answering yes/no decomposed questions. This contrasts with the separate API call approach, which more closely resembled  human response distributions, suggesting an underlying bias effect in batch prompts.

\textbf{Task-specific AES holistic rubrics show significant improvement when reducing examples, but general AES holistic rubrics do not.} Reducing the number of examples from the full set to three (3ex) tends to decrease $\tau$ for GPT on AES for the general holistic rubrics in essay prompts 1 and 4 ($\tau$ dropped from 0.437 to 0.387 and 0.695 to 0.687, respectively), and increase $\tau$ for the task-specific holistic rubric in essay prompt 6 ($\tau$ rose from 0.629 to 0.644). This could also be because the examples have explanations for essay prompts 1 and 4, but not for 6. The score explanations could be more influential than the scores alone when examples were originally provided, but this could also be due to the type of rubric.

\textbf{Adding examples for IF holistic rubrics show improvement.}
Although there is an increase in $\tau$ when adding examples to the holistic rubric in IF for both autoraters (0.536 to 0.585 and 0.470 to 0.578 for GPT and Llama respectively), it is not statistically significant.

\subsection{Decomposition Level}

Analytic rubrics do not consistently outperform holistic rubrics in aligning autoraters with human judgments due to prompt complexity or aggregation methods. 

\textbf{Prompt complexity moderates effects. }In Table~\ref{tab:asap_results} for AES, under the $\Delta$rater condition for essay prompt 1, almost all analytic batch prompts had higher correlation with humans ($\tau$ of 0.464, 0.455, 0.448, 0.439, 0.388) than the full holistic prompt ($\tau$ of 0.437). Conversely in essay prompts 4 and 6, the full holistic prompt outperformed most analytic batch prompts ($\tau$ of 0.695 compared to 0.696, 0.677, 0.652, 0.595 and $\tau$ of 0.629 compared to 0.652, 0.608, 0.557, 0.510). This discrepancy may stem from the complexity of the holistic prompts. Essay prompt 1's holistic prompt is highly complex, involving multiple sub-criteria that contain complex decisions, whereas, essay prompts 4 and 6 have less complex prompts. Introducing analytic rubrics may increase evaluation complexity, which leads to lower $\tau$. 

\textbf{Aggregation methods influence agreement. }Another factor is the prompt's output. The Pareto dominance aggregation method, a conservative estimate of essay comparison, is highly sensitive to disagreements in any single sub-criteria. In IF, both GPT-4o and Llama performed worse with analytic prompts than with holistic prompts ($\tau$ of 0.167 to 0.536 and 0.166 to 0.470). This is surprising, given that IF analytic prompts are considered task-specific, which should provide more detailed information about the task, potentially leading to higher agreement. This may be due to the aggregation method used--the ratio of ``yes'' to ``no'' responses--which does not account for the varying weights of sub-criteria in the overall holistic evaluation. Therefore, holistic preferences, being more straightforward to calculate, may yield higher performance. Higher human-autorater agreement is not necessarily achieved by decomposing a holistic rubric into several analytic parts, rather it is more important to understand the complexity of each evaluation measure as well as the aggregation methods used.

\subsection{Agreement Level}
Both datasets are stratified by human agreement level since human inter-rater agreement may influence human-autorater agreement.

\paragraph{High human inter-rater agreement is important.} On IF holistic rubrics, human-autorater agreement with the consolidated score was significantly higher when all three human annotators agreed under all conditions, than if only two or no annotators agreed with each other. This held for IF analytic rubrics for both the separate and edited conditions, although the batch condition had an increase, it was not significant. For AES holistic rubrics, a significant increase was only observed for prompt 4, while a non-significant increase was observed for prompt 6. Detailed analyses can be found in Appendix~\ref{sec:agreement}.

%% file: sections/conclusion.tex
\section{Conclusion}
This work highlights the importance of understanding the domain, autorater, and rubric during rubric creation. First, adding examples and context significantly improves human-autorater agreement in addition to autorater self-agreement, but this is dependent on the domain and autorater.
Second, rubrics that reduce confirmation bias tend to provide significant improvement for human-autorater agreement. 
Third, rubric complexity and aggregation methods across holistic and analytic rubrics influence human-autorater agreement.
Lastly, higher human inter-rater agreement contributes positively to human-autorater agreement. Practitioners aiming to use autoraters should carefully curate human annotation data and design rubrics that appropriately address the differences across domains and autoraters. Future work should explore a wider range of domains, autoraters, and rubrics to develop more comprehensive recommendations.

%% file: sections/appendix/limitations.tex
\section{Limitations}
First, the findings are drawn from only two evaluation domains (automatic essay scoring and instruction following) and two autoraters (GPT-4o and Llama-3.1-70B-Instruct). While these were chosen to provide complementary evidence, the extent to which the observed patterns generalize to other evaluation tasks (e.g., summarization, dialogue quality) or other autoraters remains unexplored.

Second, all human annotations are drawn from existing datasets rather than collected under controlled experimental conditions for this study. This means that variability in annotator training, expertise, and scoring context across datasets could influence the observed agreement patterns. Moreover the rubric modifications explored represent a small subset of possible edits. Other modifications were not tested and may interact differently with human-autorater agreement.
 
Finally, this work limited to high-resource english-language evaluation tasks, rubric sensitivity may differ across languages but is understudied, particularly for autoraters, whose performance can vary substantially by language.

%% file: sections/appendix.tex
\section{Appendix A}
\label{sec:appendix_A}
\input{sections/appendix/other_obs}
\input{sections/appendix/agreement_level}

\section{Appendix B}
\label{sec:appendix_B}
\input{sections/appendix/aes_prompts}

\input{sections/appendix/if_prompts}

%% file: sections/appendix/other_obs.tex
\subsection{Additional Other Observations}
\label{sec:other_obs}

\begin{table}[]
    \small
    \centering
    \begin{tabular}{|w{c}{0.05cm}|w{c}{0.3cm}||c|c||c|c|c|}
    \hline
        P & C. & Full & 3ex & Sep. & Bat. & Edited \\
        \hline
        \multirow{6}{*}{1}& H. & $0.639$ & $0.623$ & -& -& -\\
        \cline{2-7}
        & Ide. & -& -& $0.647$ & $0.652$ & $0.660$\\
        \cline{2-7}
        & Or. & -& -& $0.675$ & $0.626$ & $0.672$\\
        \cline{2-7}
        & WC & -& -& $0.616$ & $0.580$ & $0.639$\\
        \cline{2-7}
        & SF & -& -& $0.608$ & $0.586$ & $0.632$\\
        \cline{2-7}
        & Cv. & -& -& $0.580$ & $0.559$ & $0.610$\\
        \hline
        \multirow{6}{*}{4}& H. & $0.825$& $0.812$ & -& -& -\\
        \cline{2-7}
        & Ct. & -& -& $0.777$ & $0.815$ & $0.801$\\
        \cline{2-7}
        & PA & -& -& $0.767$ & $0.773$ & $0.784$\\
        \cline{2-7}
        & La. & -& -& $0.755$ & $0.711$ & $0.766$\\
        \cline{2-7}
        & Na. & -& -& $0.767$& $0.747$ & $0.789$\\
        \hline
        \multirow{6}{*}{6}& H. & $0.790$ & $0.782$ & -& -& -\\
        \cline{2-7}
        & Ct. & -& -& $0.716$ & $0.779$ & $0.772$\\
        \cline{2-7}
        & PA & -& -& $0.712$ & $0.746$ & $0.780$\\
        \cline{2-7}
        & La. & -& -& $0.745$ & $0.705$ & $0.753$\\
        \cline{2-7}
        & Na. & -& -& $0.749$ & $0.709$ & $0.772$\\
        \hline
    \end{tabular}
    \caption{Kendall's $\tau$ between GPT and Llama. All abbreviations follow Table~\ref{tab:asap_results}.}
    \label{tab:kendalls_twomodels}
\end{table}

\begin{figure}[hbt!]
    \includegraphics[width=\linewidth]{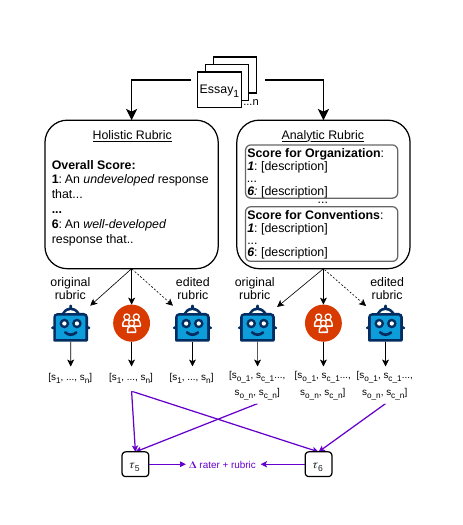}
    \vspace{-1cm}
    \caption{This diagram represents comparisons made between human-autorater agreements $\tau$ across various conditions such as holistic rubrics (left side), in which all criteria are applied together in a single overall judgment, or analytic rubrics (right side), in which criteria are evaluated separately, and rubrics which have been edited for LLMs. Arrow in bold between $\tau_5$ and $\tau_6$ represent comparisons for which statistical significance can be calculated.}
    \label{fig:experiment_design_app}
\end{figure}

\begin{table*}[h!]
    \small
    \centering
    \begin{tabular}{|c|c||c|c||c|c|c||c|c||c|c|c|}
    \hline
        & & \multicolumn{10}{c|}{$\Delta\textbf{rater+rubric}$} \\
        \hline
        & & \multicolumn{5}{c||}{GPT-4o} & \multicolumn{5}{c|}{Llama}\\
        \hline
        & & \multicolumn{2}{c||}{$\textnormal{LLM}_{H} \to \textnormal{H}_A$}& \multicolumn{3}{c||}{$\textnormal{H}_H \to \textnormal{LLM}_A$} & \multicolumn{2}{c||}{$\textnormal{LLM}_{H} \to \textnormal{H}_A$}& \multicolumn{3}{c|}{$\textnormal{H}_H \to \textnormal{LLM}_A$}\\
        \hline
        \hline
        \multirow{3}{*}{\rotatebox[origin=c]{90}{AES}} & 1 & $0.574^{\dagger}$& $0.539$& $0.279$ &$0.271$ & $0.401^{s,b\uparrow}$ & $0.601$& $0.605$& $0.467$ & $0.467$&$0.470$\\
        \cline{2-12}
        & 4 & $0.691$& $0.688$& $0.669$ &$0.668$ &$0.682^{s,b\uparrow}$& $0.674$& $0.674$& $0.710^{\star}$ & $0.674$ & $0.707^{b\uparrow}$\\
        \cline{2-12}
        & 6 & $0.645$ & $0.653^{\dagger}$& $0.585$ &$0.595^{\star}$ & $0.644^{s,b\uparrow}$ & $0.660$& $0.670^{\dagger}$& $0.637^{\star}$ &$0.592$ & $0.671^{s,b\uparrow}$\\
        \hline
        \hline
        IF & & $0.541$& $0.530$&  $0.474^{\star}$& $0.34$& $0.459 ^{b\uparrow}$ & $0.478$& $0.545$&  $0.421^{\star}$& $0.183$& $0.424^{b\uparrow}$\\
        \hline
    \end{tabular}
    \caption{Kendall's $\tau$ results for $\Delta$rater+rubric with GPT-4o and Llama on AES and IF. For AES, the number indicates the essay prompt, with ideas, organization, word choice, sentence fluency, and conventions compared for prompt 1, and content, prompt adherence, language, and narrativity compared for prompts 4 and 6. $\tau$ is calculated through Pareto dominance comparison for AES and through instruction following ratio comparison for IF. Significance tests between separate (sep.), batch (bat.), and edited prompts are performed, where ${^s}$ and ${^b}$ in the edited prompt column represents significant differences with separate and batch prompts respectively. $\dagger$ is indicated next to comparisons that are significantly larger within holistic prompts. $\star$ is indicated next to comparisons that are significantly larger between separate and batch comparisons. $\uparrow$ and $\downarrow$ represent that the $\tau$ value for edited prompts is significantly larger or smaller respectively with the separate $s$ or batch $b$ prompts' $\tau$. The lack of any dagger, star, or arrow denotes no statistical significance. $\textnormal{H}$ is a shortened form for $\textnormal{Human}$.}
    \label{tab:rater_rubric_combined}
\end{table*}

\paragraph{$\Delta$rater + rubric.}Different raters may show greater agreement when using different rubrics. In additional to the previous comparisons of \textbf{$\Delta$Rater} and \textbf{$\Delta$Rubric} we also investigate \textbf{$\Delta$Rater+Rubric}, which captures cases where alignment between different raters is affected by the use of different rubric types as depicted in Figure \ref{fig:experiment_design_app}. The best $\tau$ obtained for any rubric condition when comparing human ratings on holistic rubrics with autoraters on analytic rubrics ($\textnormal{Human}_{H}$ $\to$ $\textnormal{LLM}_{A}$, 0.401, 0.682, and 0.644 respectively) is lower than the worst $\tau$ obtained from comparing autoraters on holistic rubrics with humans on analytic rubrics ($\textnormal{LLM}_{H}$ $\to$ $\textnormal{Human}_{A}$, 0.539, 0.688, and 0.645 respectively) (Table~\ref{tab:asap_results}). This indicates that while autoraters exhibit high preference agreement across different rubrics (in $\Delta$ rubric), the Pareto dominance aggregation method decreases alignment between autoraters and humans. This trend is also observed in Table~\ref{tab:if_results} (0.474 to 0.530, and 0.424 to 0.478) for IF, suggesting its applicability to the ratio aggregation method. However, in Table~\ref{tab:asap_results} on Llama for AES, this only occurs for essay prompt 1 (0.470 to 0.601).

Surprisingly, for essay prompt 4, all $\tau$ values for $\Delta$rater + rubric are higher than those for $\Delta$ rubric when humans are the annotator, for both GPT-4o (0.656 compared with 0.691 and 0.688) and Llama (0.656 compared with 0.674 and 0.674). This suggests that the autorater's preferences on the holistic rubric agree slightly more than the human's preferences on the holistic rubric as compared to the analytic rubric. Although different humans rate with the holistic and analytic rubrics, they all either were trained or familiar with evaluation, whereas there is no guarantee for autorater performance. This could be due to various factors, such as the autorater using a continuous scale versus the human's discrete scale which leads to less ties, or the autorater's interpretation being closer to the human's analytic rubric interpretation. Future work should further explore these correlations.

\paragraph{GPT-4 vs. Llama.}It is hypothesized that autoraters will exhibit greater mutual alignment  when (1) using analytic prompts instead of holistic prompts, (2) using the same (and more) examples, and (3) using separate API calls instead of batched API calls. To test these hypotheses, Kendall's $\tau$ is calculated between GPT-4 ratings and Llama ratings in Table~\ref{tab:kendalls_twomodels}. Analytic rubrics do not consistently improve autorater alignment over holistic rubrics, even when provided examples in AES (0.639 compared to 0.610 - 0.672, 0.825 compared to 0.766 - 0.801, and 0.790 to 0.753 - 0.780 respectively). However, introducing examples to analytic rubrics (comparing separate and batch with the edited column) and increasing the number of examples in holistic rubrics (0.639 from 0.623, 0.825 from 0.812, and 0.790 from 0.782) does tend to improve alignment, indicating that examples improve both human-autorater and autorater-autorater alignment in AES. This pattern does not hold for IF, where the separate, batch, and edited conditions yield $\tau$ values of 0.674, 0.193, and 0.662, respectively, while the holistic zero-shot and three-shot conditions yield 0.752 and 0.724. Separate API calls improve alignment for 8 out of 13 sub-criteria in AES. However, as previously mentioned, further research is needed to understand  how humans and autoraters exhibit positional bias, as the sub-criteria presented first in batched API calls consistently show higher alignment.

%% file: sections/appendix/agreement_level.tex
\begin{table*}[]
    \centering
    \small
    \begin{tabular}{c >{\RaggedRight\arraybackslash}p{6cm} l >{\RaggedRight\arraybackslash}p{5cm}}
    \toprule
        \textbf{Prompt} & \textbf{Source} & \textbf{Rubric Type} & \textbf{Distinction} \\
        \midrule
        1 & Argumentative (students asked to state an opinion based on the prompt) & General & Evaluates opinion with a reusable rubric \\
        \midrule
        4 & Source-dependent (students asked to read a passage and respond to a prompt with details from the passage) & General & Evaluates comprehension of a source using a reusable rubric \\
        \midrule
        6 & Source-dependent (students asked to read a passage and respond to a prompt with details from the passage) & Task-specific & Evaluates comprehension using a highly specific, non-reusable rubric \\
        \bottomrule
    \end{tabular}
    \caption{ASAP Chosen Essay Prompts}
    \label{tab:asap_prompts}
\end{table*}

\subsection{Agreement Level}
\label{sec:agreement}
\begin{table}[ht!]
    \centering
    \begin{tabular}{c|cccc}
        \toprule
        \textbf{M} & \textbf{Prompt} & \textbf{Comp.} & \textbf{Full} & \textbf{3ex} \\
        \midrule
        \multirow{6}{*}{\rotatebox[origin=c]{90}{GPT-4o}} & \multirow{2}{*}{1} & agree & $0.450$ & $0.398$ \\
        & & disagree & $0.454$ & $0.413$ \\
        \cline{2-5}
        & \multirow{2}{*}{4} & agree & $0.687^{\dagger}$ & $0.678^{\dagger}$ \\
        & & disagree & $0.525$ & $0.512$ \\
        \cline{2-5}
        & \multirow{2}{*}{6} & agree & $0.659$ & $0.675$ \\
        & & disagree & $0.642$ & $0.653$ \\
        \hline\hline
        \multirow{6}{*}{\rotatebox[origin=c]{90}{Llama}} & \multirow{2}{*}{1} & agree & $0.585$ & $0.581$\\
        & & disagree & $0.593$ & $0.592$ \\
        \cline{2-5}
        & \multirow{2}{*}{4} & agree & $0.694^\dagger$ & $0.691^\dagger$ \\
        & & disagree & $0.548$ & $0.529$ \\
        \cline{2-5}
        & \multirow{2}{*}{6} & agree & $0.696$ & $0.707$ \\
        & & disagree & $0.677$ & $0.693$ \\
        \bottomrule
    \end{tabular}
    \caption{Kendall's $\tau$ results on AES with GPT-4o and Llama. The number indicated by the $\dagger$ represents that the $\tau$ value on that subset of data (in all cases here, where raters agreed with each other) is significantly larger than the $\tau$ value on the other subset of data (where raters disagreed with each other). The lack of any arrow denotes no statistical significance.}
    \label{tab:tau_aes_gpt_agree}
\end{table}

\begin{table*}[h!]
    \centering
    \begin{tabular}{c|c c c c c c}
        \toprule
        \textbf{M} & \textbf{Comp.} & \textbf{0ex} & \textbf{3ex} & \textbf{Separate} & \textbf{Batch} & \textbf{Edited} \\
        \midrule
        \multirow{3}{*}{\rotatebox[origin=c]{90}{GPT-4o}} &Full agreement & $0.792^{p\uparrow, f\uparrow}$ & $0.803^{p\uparrow, f\uparrow}$ & $0.299^{p\uparrow}$ & $0.078$ & $0.31^{p\uparrow}$ \\
        &Partial disagreement & $0.559^{f\uparrow}$ & $0.584^{f\uparrow}$ & $0.251$ & $0.028$ & $0.261$ \\
        &Full disagreement & $0.220$ & $0.268$ & - & - & - \\
        \hline\hline
        \multirow{3}{*}{\rotatebox[origin=c]{90}{Llama}}&Full agreement & $0.730^{p\uparrow,f\uparrow}$& $0.865^{p\uparrow,f\uparrow}$ & $0.283^{p\uparrow}$& $0.132$ & $0.265^{p\uparrow}$ \\
        &Partial disagreement & $0.533^{f\uparrow}$& $0.625^{f\uparrow}$&$0.242$ &$0.017$ &$0.203$ \\
        &Full Disagree& $0.083$& $0.229$& -& -&- \\
        \bottomrule
    \end{tabular}
    \caption{Kendall's $\tau$ results on IF with GPT-4o for $\Delta$rater. Significance tests between data with full agreement, partial agreement, and full disagreement are performed, where $^{p}$ and $^{f}$ in the full agreement row represents significant differences with partial disagreement and full disagreement data respectively. $\uparrow$ and $\downarrow$ represent that the $\tau$ value is significantly larger or smaller respectively with compared $\tau$. The lack of any arrow denotes no statistical significance.}
    \label{tab:tau_if_gpt_agree}
\end{table*}

Agreement level is examined using holistic prompts in both AES and IF, and analytic prompts in IF. As shown in Table~\ref{tab:tau_aes_gpt_agree} with GPT-4o, for essay prompts 4 and 6 (where both raters agreed on 77.1\% and 62.2\% of essays, respectively), essays with rater agreement exhibited higher $\tau$ than those without (for full examples, $\tau$ of 0.687 to 0.525 and 0.659 to 0.642 respectively, for 3ex, $\tau$ of 0.678 to 0.512 and 0.675 to 0.653 respectively), with a significant difference in essay prompt 4. However, this pattern did not hold for essay prompt 1 (where raters agreed on 65.3\% of essays, for full examples, $\tau$ rose from 0.450 to 0.454, and for 3ex, $\tau$ rose from 0.398 to 0.413 albeit not significantly). Despite greater disagreement among human raters, the autorater showed higher agreement with the aggregated human scores. This discrepancy might stem from the different aggregation methods used: The final score for essay prompt 1 is the average of both raters' scores, while essay prompts 4 and 6 use a single rater's score (or a third expert rater's score). Averaging scores could diminish the impact of rater disagreement on $\tau$ in essay prompt 1. Llama ratings mirror these patterns.

In the IF dataset, holistic prompt ratings (with three human raters) were divided into three subsets: full agreement (all three raters agree, 25.9\% of the dataset), partial agreement (two raters agree, 53.3\% of the dataset), and full disagreement (no raters agree, 20.9\% of the dataset). The final score was obtained by averaging the three raters' scores. Analytic prompt ratings were divided into full agreement (13\% of the dataset) and partial agreement (87\% of the dataset) subsets, using majority vote for the final score. Kendall's $\tau$ was calculated within the subsets for each prompt. Table~\ref{tab:tau_if_gpt_agree} demonstrates a consistent trend for both GPT-4o and Llama: lower human agreement corresponded to lower $\tau$, with most comparisons showing statistical significance (for GPT-4o, for the holistic prompt, with no examples, $\tau$ dropped from 0.792 to 0.559 to 0.220, with 3 examples, $\tau$ dropped from 0.803 to 0.584 to 0.268; while for the analytic prompt, separate dropped from 0.299 to 0.251, batch from 0.078 to 0.028, and edited from 0.31 to 0.261 respectively) . To achieve high human-autorater alignment, one must first establish high human-human alignment. While a rubric cannot fix inherently ambiguous data, it is the primary tool for reducing rater disagreement and creating the stable ‘ground truth’ necessary for meaningfully evaluating and training an autorater.

%% file: sections/appendix/aes_prompts.tex

\subsection{Prompt for AES}
\label{sec:aes_prompt_appendix}
The holistic and analytic rubrics within the prompts are taken from ASAP \citep{asap-aes} and ASAP++ \citep{mathias-bhattacharyya-2018-asap}. The contextual information present in the grading guidelines are formatted into the prompt, and additionally added with slight edits to the analytic prompts in the edited condition. The examples chosen from ASAP++ are essay IDs 449, 1264, 1616, 9125, 9430, 9497, 15558, 16520, and 16581.

\begin{tcolorbox}[title=Prompt 1 Holistic Context,breakable]
Instructions:

1) The following essay is a first draft written by an 8th grade student in forty-five minutes in reaction to a prompt designed to elicit persuasive writing. You will score these timed responses holistically, which means that you will determine a score based on the overall impression most often gained from a single reading of the response.

2) You will be given a rubric that outlines a six-point scale. Each score point on that scale is described by an overall statement which captures the essence of the response. The elements of the response (elaboration, organization, fluency and audience awareness) that are typical for that score point are described below the overall statement. Individual responses may be stronger in one feature and weaker in another. In other words, the list of features at each score point, while helpful, cannot perfectly describe every response in a score point category.

3) A committee of expert readers uses this rubric as a guide to select anchor papers for each score point. Anchor papers are examples of actual student work. The committee prepares an anchor set composed of several papers at each score point. They deliberately select papers to show an appropriate range of writing skill for each score point and to represent the variety of approaches students take when addressing the writing prompt. You rely heavily on these anchor sets to guide your scoring.

4) Errors in spelling, punctuation, grammar, and usage are not considered as part of the criteria for scoring. Also, papers receive a score based on the work the student did complete even if they seem to be unfinished. Because the writing sample is a timed response, it is generally assumed that these errors and omissions could have been corrected if the student had been given an opportunity to revise and edit. You are trained to read through these errors when you score student papers.

5) Score the essay on a scale from 1 to 6.
\end{tcolorbox}

\begin{tcolorbox}[title=Prompt 4 Holistic Context,breakable]
    Instructions:
    
1) The following essay is written by an 10th grade student in response to a prompt that is dependent on reading the story provided.

2) You will be given a rubric that outlines a four-point scale.

3) Training materials consist of a rubric and a scoring guide of ten responses.

4) Score the essay on a scale from 0 to 3.
\end{tcolorbox}

\begin{tcolorbox}[title=Prompt 6 Holistic Context,breakable]
    Instructions:
    
1) The following essay is written by an 10th grade student in response to a prompt that is dependent on reading the excerpt provided.

2) You have a four year baccalaureate degree as well as documented coursework in English. You are not a teacher, substitute teacher, support staff, tutor, administrator, etc., who is currently under contract or employed by or in schools, or under 18 years of age.

3) You will be given a rubric that outlines a five-point scale.

4) You will be given an anchor set which will consist of responses that are typical, rather than unusual or uncommon; solid, rather than controversial or borderline; and true, meaning that these have scores that cannot be changed by anyone other than pertinent personnel. Anchor sets will typically have 2 to 3 sample responses at each score point (the middle score points will have 3 sample responses, 1 representing the mid-high to high end of the score point range, 1 in the middle, and 1 at the mid-low to low end).

5) Score the essay on a scale from 0 to 4.
\end{tcolorbox}

\begin{tcolorbox}[title=Example of context given to an analytic prompt for separate (Prompt 1),breakable]
Instructions:

1) The following essay is a first draft written by an 8th grade student in forty-five minutes in reaction to a prompt designed to elicit persuasive writing.

2) You will be given a rubric that outlines a six-point scale for an attribute. 

3) Score the essay on a scale from 1 to 6 on the attribute.
\end{tcolorbox}

\begin{tcolorbox}[title=Example of context given to an analytic prompt for edited (Prompt 1),breakable]
Instructions:

1) The following essay is a first draft written by an 8th grade student in forty-five minutes in reaction to a prompt designed to elicit persuasive writing.

2) You will be given a rubric that outlines a six-point scale for an attribute. 

3) A committee of expert readers uses this rubric as a guide to select anchor papers for some score points. Anchor papers are examples of actual student work. The committee prepares an anchor set composed of several papers at various score points. They deliberately select papers to show an appropriate range of writing skill and to represent the variety of approaches students take when addressing the writing prompt. You rely heavily on these anchor sets to guide your scoring.

4) Also, papers receive a score based on the work the student did complete even if they seem to be unfinished. Because the writing sample is a timed response, it is generally assumed that these errors and omissions could have been corrected if the student had been given an opportunity to revise and edit. You are trained to read through these errors when you score student papers.

5) Score the essay on a scale from 1 to 6 on the attribute.
\end{tcolorbox}

%% file: sections/appendix/if_prompts.tex
\subsection{Prompt for Instruction Following}
\label{sec:if_prompt_appendix}
The holistic and analytic prompt are structured to match the annotation instructions presented to human annotators of the InfoBench dataset \citep{qin-etal-2024-infobench}.

\begin{tcolorbox}[title=Evaluation Prompt,breakable]
    I will show you examples of how to evaluate system responses to specific criteria.

EXAMPLE 1: \{\}

EXAMPLE 2: \{\}

EXAMPLE 3: \{\}

Now, please evaluate the current response based on the following criteria:

Based on the provided Input and Generated Text, answer the following Question with either a YES or NO choice. Your selection should be based on your judgment as well as the following rules: 

- YES: Select 'YES' if the generated text entirely fulfills the condition specified in the question. However, note that even minor inaccuracies exclude the text from receiving a 'YES' rating. As an illustration, consider a question that asks, "Does each sentence in the generated text use a second person?" If even one sentence does not use the second person, the answer should NOT be 'YES'. To qualify for a 'YES' rating, the generated text must be entirely accurate and relevant to the question. 

- NO: Opt for 'NO' if the generated text fails to meet the question's requirements or provides no information that could be utilized to answer the question. For instance, if the question asks, "Is the second sentence in the generated text a compound sentence?" and the generated text only has one sentence, it offers no relevant information to answer the question. Consequently, the answer should be 'NO'.

Input: \{\}
Generated Text (System's Response): \{\}

Question: Does every chapter in the generated list include a description?
\end{tcolorbox}



\begin{tcolorbox}[title=Example 1,breakable]
Instruction: Create two tables with three columns each, using \"|\" to separate the columns, and include an appropriate header for each column:\\a. In the first table, display seven interesting geological formations or landmarks worldwide, including a short description and location. Limit descriptions to 40 words.\\b. In the second table, present three main types of rocks, a brief description of their formation process, and common examples. Limit each rock type description to 50 words.\\Generated Text (System's Response): a. Geological Formations/Landmarks Table:\\\\| Formation Name | Description | Location |\\|----------------|-------------|----------|\\| Grand Canyon | A vast, colorful canyon with layered bands of red rock revealing millions of years of geological history. | Arizona, USA |\\| Great Barrier Reef | The world's largest coral reef system, visible from space, teeming with marine life. | Queensland, Australia |\\| Mount Everest | Earth's highest mountain above sea level, a part of the Himalayan range. | Nepal/China border |\\| Dead Sea | A salt lake whose shores are the lowest land-based elevation on Earth, known for its buoyancy. | Jordan/Israel border |\\| Uluru (Ayers Rock) | A massive sandstone monolith sacred to indigenous Australians, known for its red hue. | Northern Territory, Australia |\\| Yellowstone Caldera | A volcanic caldera and supervolcano with geysers, hot springs, and diverse wildlife. | Wyoming, USA |\\| Giant's Causeway | An area of about 40,000 interlocking basalt columns, the result of an ancient volcanic eruption. | County Antrim, N. Ireland |\\\\b. Rock Types Table:\\\\| Rock Type | Formation Process | Common Examples |\\|-----------|-------------------|-----------------|\\| Igneous | Formed from the solidification of molten rock material (magma or lava). | Granite, basalt, obsidian |\\| Sedimentary | Created by the accumulation and compaction of sediment or by the precipitation of minerals from water. | Sandstone, limestone, shale |\\| Metamorphic | Formed through the alteration of other rocks under high pressure and temperature without melting. | Slate, marble, gneiss |\\
\{Questions for Example 1\}
\end{tcolorbox}

\begin{tcolorbox}[title=Questions for Example 1,breakable]
Question 1: Does the generated text include exactly two tables?\\Answer: YES\\Question 2: Do both tables in the generated text consist of three columns each?\\Answer: YES\\Question 3: Are the columns in both tables in the generated text separated by \"|\"?\\Answer: YES\\Question 4: Do both tables in the generated text include an appropriate header for each column?\\Answer: YES\\Question 5: Does the first table in the generated text display seven interesting geological formations or landmarks worldwide?\\Answer: YES\\Question 6: Does each entry in the first table of the generated text include a short description and location for each geological formation or landmark?\\Answer: YES\\Question 7: Are the descriptions in the first table of the generated text limited to 40 words or less for each geological formation or landmark?\\Answer: YES\\Question 8: Does the second table in the generated text present exactly three main types of rocks?\\Answer: YES\\Question 9: Does each entry in the second table in the generated text include a brief description of the rock type's formation process?\\Answer: YES\\Question 10: Does each entry in the second table in the generated text provide common examples for each rock type?\\Answer: YES\\Question 11: Are the descriptions in the second table of the generated text limited to 50 words or less for each rock type?\\Answer: YES\\
\end{tcolorbox}

\begin{tcolorbox}[title=Example 2,breakable]
Input: A research study has been conducted to determine if exercise really can "boost" your metabolism.
\\Instruction: You should choose a YouTube video title based on the video's content. A video's title tells viewers what to expect from it. It should be direct, honest, and clear. The title of the video needs to capture the attention of viewers, so do not use an unclear or ambiguous one.
\\Generated Text (System's Response): Yes, exercise can increase your metabolism. Studies have found that when you exercise, your body produces more heat and requires more energy to perform basic tasks. This can lead to an increase in metabolic rate of up to 14\%, which can last up to 13 hours after your workout. Additionally, regular exercise can also help to reduce fatigue and increase energy levels.\\Question 1: Is the generated text a YouTube video title?\\Answer: NO\\Question 2: Is the generated title direct, honest, and clear?\\Answer: NO\\Question 3: Is the generated title unambiguous?\\Answer: NO\\Question 4: Is the generated title based on the given video content?\\Answer: NO\\Question 5: Does the generated title tell viewers what to expect?\\Answer: NO\\
\end{tcolorbox}

\begin{tcolorbox}[title=Example 3,breakable]
Input: Item, Qty, Price, Subtotal | mango, 5, 500, ? | apple, 4, 150, ? | oil, 2, 1500, ?  \\Function: column2 * column 3\\Instruction: We have entered the home supplies budget in the following table. Calculate the last column with the given function and rewrite the completed table.\\Note that, columns in a row are separated using a comma, and rows are separated using a "|".\\Generated Text (System's Response): Item, Qty, Price, Subtotal \\mango, 5, 500, 2500 \\apple, 4, 150, 600 \\oil, 2, 1500, 3000\\Question 1: Is the generated text a table?\\Answer: NO\\Question 2: Are the columns in a row in the generated table separated by a comma?\\Answer: YES\\Question 3: Are the rows in the generated table separated by a \"|\"?\\Answer: NO\\Question 4: Is the generated table complete?\\Answer: YES\\Question 5: Is the generated table copied from the given table except for the last column?\\Answer: NO\\Question 6: Is the last column in the generated table correctly calculated by the function in the given input?\\Answer: YES\\
\end{tcolorbox}